\PassOptionsToPackage{table}{xcolor}
\documentclass{applemlr}
\usepackage{latexsym}
\usepackage{listings}
\usepackage{array}
\usepackage{amsmath}
\usepackage{tabularx}
\usepackage{amssymb}
\usepackage{wrapfig}
\usepackage{needspace}
\usepackage{enumitem}
\usepackage{pifont}
\usepackage{bbding}
\usepackage{float}
\usepackage{makecell}

\graphicspath{{figures/}}
\lstdefinestyle{promptstyle}{
  basicstyle=\normalsize\rmfamily,
  breaklines=true,
  breakatwhitespace=false,
  columns=fullflexible,
  keepspaces=true,
  tabsize=2,
  aboveskip=0pt,
  belowskip=0pt,
  xleftmargin=0pt,
  framexleftmargin=0pt,
}

\newtcolorbox{promptbox}[1][]{%
  colback=gray!5,
  colframe=gray!50,
  fonttitle=\bfseries\small\rmfamily,
  title={#1},
  enhanced,
  left=3pt, right=3pt, top=4pt, bottom=4pt,
}

\newtcolorbox{casebox}[1][]{%
  colback=gray!5,
  colframe=gray!55,
  colbacktitle=gray!22,
  coltitle=black,
  fonttitle=\bfseries\small\rmfamily,
  title={#1},
  enhanced,
  arc=1.5mm,
  left=6pt, right=6pt, top=4pt, bottom=4pt,
  boxsep=2pt,
}

\title{WorldLines: Benchmarking and Modeling Long-Horizon Stateful Embodied Agents}

\author[1,3\dagger]{Yehang Zhang}
\author[1,3\dagger]{Jianchong Su}
\author[1,3\dagger]{Haojian Huang}
\author[3]{Yifan Chang}
\author[1,3]{Tianhao Zhou}
\author[1]{Xinli Xu}
\author[1,3]{Yingjie Xu}
\author[3]{Yinchuan Li}
\author[3\ddagger]{Zexi Li}
\author[1,2\ast]{Ying-Cong Chen}

\affiliation[1]{HKUST(GZ)}
\affiliation[2]{HKUST}
\affiliation[3]{Knowin}
\contribution[\dagger]{Equal Contribution}
\contribution[\ddagger]{Project Leader}
\contribution[\ast]{Corresponding Author}

\abstract{To assist humans over extended periods in real homes, embodied agents must remember user routines, world states, and past interactions. Existing long-term memory benchmarks mainly evaluate language-centric retrieval and question answering, while embodied benchmarks often focus on short-horizon task execution without testing long-term memory use in dynamic environments. We introduce WorldLines, a project-driven benchmark for long-horizon embodied household assistance. It constructs temporally extended household traces with dialogues, actions, execution feedback, object and device state changes, and converts them into evidence-linked samples for Memory QA and Embodied Task Planning. We further propose ObsMem, an observer-grounded memory framework that maintains visibility-aware memories and action-native state trails for state-aware decisions. Experiments reveal persistent challenges in partial observability, overwritten world states, and translating long-term memory into embodied plans, while ObsMem offers a stronger reference architecture for this setting.}

\begin{document}
\maketitle

\section{Introduction}

To operate reliably over long horizons, embodied agents need more than memory of past interactions; they must maintain a stateful view of an evolving world \cite{fung2025embodiedaiagentsmodeling, yang2025embodiedbench, chu2026agenticworldmodelingfoundations}.
\begin{wrapfigure}{r}{0.40\textwidth}
\vspace{-0.2em}
\centering
\includegraphics[width=\linewidth]{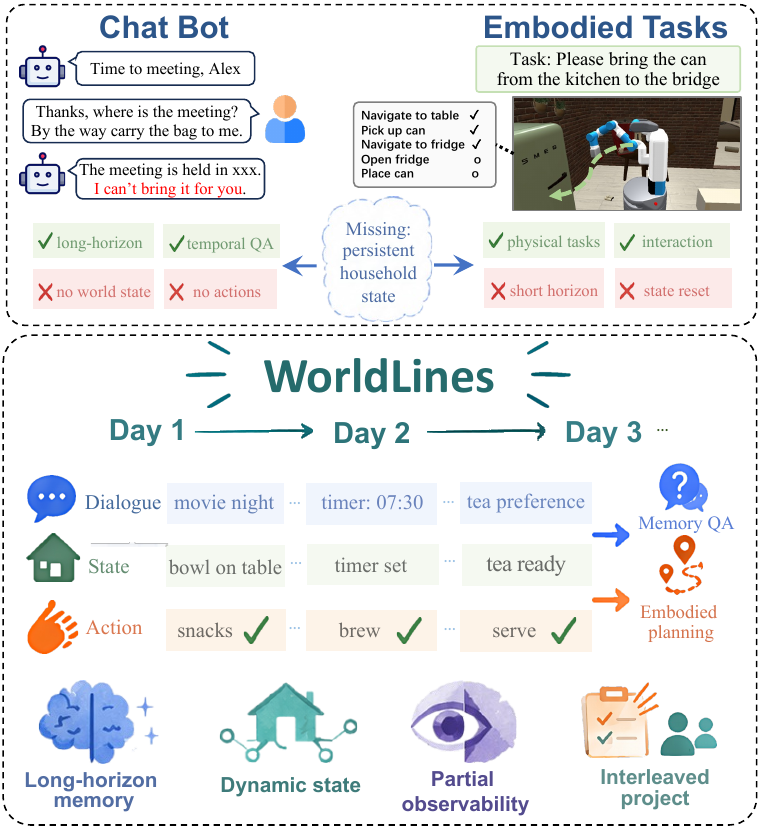}
\vspace{-1.9em}
\caption{
\textbf{Overview of WorldLines.}
WorldLines tracks cross-day dialogue, state changes, and actions for memory QA and state-aware embodied planning.
}
\vspace{-1.6em}
\label{fig:teaser}
\end{wrapfigure}
Real service requests often unfold over time and depend on user routines, object states, device settings, and recent events. For example, a user may say: ``I am going to the gym at 7:30 and will be back home at 8:30. After I return, I would like to watch a movie in the living room as usual and have something to eat. I also just bought some fruit and put it in the refrigerator.'' Responding correctly requires the robot to connect the current instruction with prior schedules, preferences, and environmental states.

This challenge becomes sharper in embodied settings. Long-horizon embodied tasks should not be reduced to isolated dialogues, single action episodes, or one-off state changes. Existing embodied benchmarks have advanced navigation, rearrangement, manipulation, and multi-agent planning~\cite{li2023behavior1k, puig2023habitat3}, but they typically remain bounded within short episodes where state does not persist across interactions. Real long-horizon interaction instead requires agents to maintain an evolving world state across dialogue, human activity, robot actions, and device changes. Because the world is partially observable, objects may be moved outside the robot's view, and container or device states may change without direct observation. Therefore, this work focuses not on single-task execution, but on whether agents can maintain partially observable world states and use them for later question answering, planning, and execution.

As summarized in Table~\ref{tab:benchmark-comparison}, existing benchmarks split this problem into two incomplete settings. Long-term memory benchmarks evaluate cross-session retrieval, updating, and question answering, but usually decouple memory from physical state transitions, action feedback, and executable constraints~\cite{wu2024longmemeval,maharana2024locomo}. Embodied benchmarks cover navigation, rearrangement, manipulation, and multi-agent planning, but are mostly confined to short episodes, where world states rarely persist across interactions or affect later tasks~\cite{chang2025partnr,shridhar2021alfworld}. This raises a central question for embodied-agent evaluation: \textit{can agents maintain persistent state over long-horizon, partially observable interactions and use it for downstream embodied tasks?}

This motivates \textbf{WorldLines}, a benchmark for evaluating long-horizon stateful embodied agents (Figure~\ref{fig:teaser}). WorldLines generates extended traces that include dialogue, human activity, robot actions, device control, execution feedback, and world-state changes, and converts them into evidence-linked Memory QA and Embodied Task Planning samples. In this setting, memory is not the endpoint of evaluation; it is the mechanism through which agents maintain state, trace evidence, and make later decisions.

WorldLines shows that long-horizon embodied agents require more than flat text-retrieval memory. Text-snippet memories \cite{xu2025amem,kang2025memoryos,chhikara2025mem0, xu2026structmemstructuredmemorylonghorizon} struggle to distinguish direct observations, reported information, and unobserved changes, and to track action-induced updates to objects, containers, and devices. We therefore introduce \textbf{ObsMem}, an observer-grounded memory framework that separates historical evidence, structured world states, and agent beliefs to support persistent state maintenance and embodied decision making under partial observability.

\begin{table*}[t]
\centering
\footnotesize
\setlength{\tabcolsep}{2pt}
\begin{tabular}{lccccc}
\toprule
\textbf{Benchmark} 
& \textbf{Setting} 
& \textbf{Long-Term} 
& \textbf{Project-Driven} 
& \textbf{Persistent World State} 
& \textbf{Physical \& Device Ops.} \\
\midrule
LongMemEval & Dialogue & \checkmark & -- & -- & -- \\
LoCoMo & Dialogue & \checkmark & -- & -- & -- \\
RealMem & Dialogue & \checkmark & \checkmark & Project state & -- \\
MEMENTO & Embodied & -- & -- & -- & \checkmark \\
PARTNR & Embodied & -- & -- & -- & \checkmark \\
\textbf{WorldLines} & \textbf{Household Sim.} & \checkmark & \checkmark & \textbf{World state} & \checkmark \\
\bottomrule
\end{tabular}
\caption{
Comparison of representative long-term memory and embodied-agent benchmarks. WorldLines combines project-driven long-term memory with persistent household world states, physical actions, and smart-device operations in simulated household environments.
}
\label{tab:benchmark-comparison}
\end{table*}

The main contributions of this work are as follows:
\begin{itemize}
    \item We introduce \textbf{WorldLines}, a benchmark for long-horizon stateful embodied agents, covering Memory QA and Embodied Task Planning in dynamic, partially observable environments.
    \item We develop a project-driven trace generation pipeline that turns grounded worlds, long-term activity threads, executable actions, and evolving states into evidence-linked evaluation samples.
    \item We propose \textbf{ObsMem}, an observer-grounded memory framework that separates event evidence, state trails, and belief records for long-horizon embodied QA and planning.
\end{itemize}

\section{Related Work}
\begin{figure*}[t]
    \centering
    \includegraphics[width=0.92\textwidth]{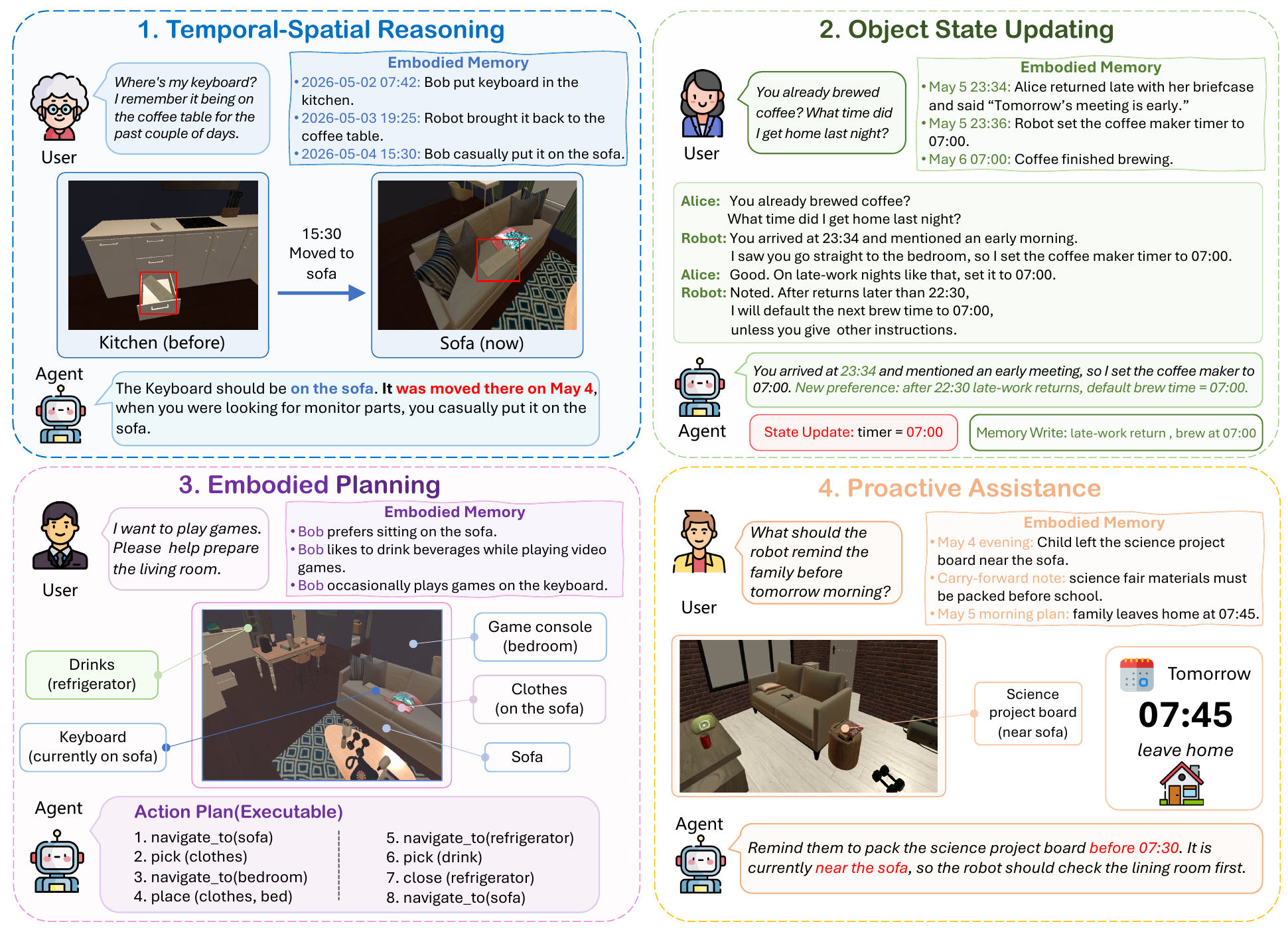}
    \caption{
    \textbf{Core dimensions of WorldLines.}
    WorldLines evaluates four aspects of long-horizon embodied tasks: temporal-spatial reasoning, object state, embodied planning, and proactive assistance.
    }
    \label{fig:task_examples}
\end{figure*}
\subsection{Benchmarks for Long-Horizon Agent Memory}
Long-term memory benchmarks for LLM-based agents have been developed primarily in conversational and multimodal settings. LoCoMo \cite{maharana2024locomo} evaluates cross-session dialogue memory, LongMemEval \cite{wu2024longmemeval} studies long-span memory updating, HaluMem \cite{chen2025halumem} focuses on memory consistency and hallucination, and RealMem \cite{bian2026realmem} introduces project-oriented long-term interaction. These benchmarks provide useful protocols for retrieval, QA, and consistency evaluation, but remain largely text-centric. Embodied benchmarks cover complementary settings: ALFWorld \cite{shridhar2021alfworld} studies language-conditioned household tasks, ProcTHOR \cite{deitke2022} and Habitat 3.0 \cite{puig2023habitat3} support simulated navigation and interaction, while PARTNR \cite{chang2025partnr} and BEHAVIOR-1K \cite{li2023behavior1k} focus on collaboration, rearrangement, and long-horizon task execution. EvoEmpirBench \cite{Zhao2025EvoEmpirBenchDS} further evaluates dynamic spatial reasoning under partial observability, but centers on game-like navigation and elimination tasks. However, they do not explicitly evaluate long-term memory in embodied task completion.
\subsection{Agent Memory Systems}
Existing LLM-based agents commonly use external memory to store, update, and retrieve information beyond a single context window \cite{hu2026memoryageaiagents}. MemGPT \cite{packer2024memgpt} organizes context and external storage into hierarchical memory tiers, while MemoryBank \cite{zhong2023memorybank} accumulates user-specific memories from long-term conversations. Recent systems further improve memory management through operating-system-inspired scheduling \cite{kang2025memoryos}, scalable extraction and update pipelines \cite{chhikara2025mem0}, agentic memory organization \cite{xu2025amem}, and graph-structured relational memory \cite{hu2026doesmemoryneedgraphs}. These works mainly study persistent memory for conversational or general-purpose agents. Memory has also been explored in embodied agents. MEMENTO \cite{kwon2025memento} studies personalized embodied assistance, while semantic-map and scene-graph methods maintain structured object, spatial, and relational knowledge for planning \cite{rana2023sayplan,gu2024conceptgraphs}. Other approaches retrieve past observations for embodied decision making \cite{xie2024embodiedrag,wang2024karma, zhou2024hazard, lillemark2026flowequivariantworldmodels} or store reusable skills and programs for future tasks \cite{wang2024voyager}.

\section{WorldLines Benchmark Construction}
\begin{figure*}[t]
    \centering
    \includegraphics[width=0.9\textwidth]{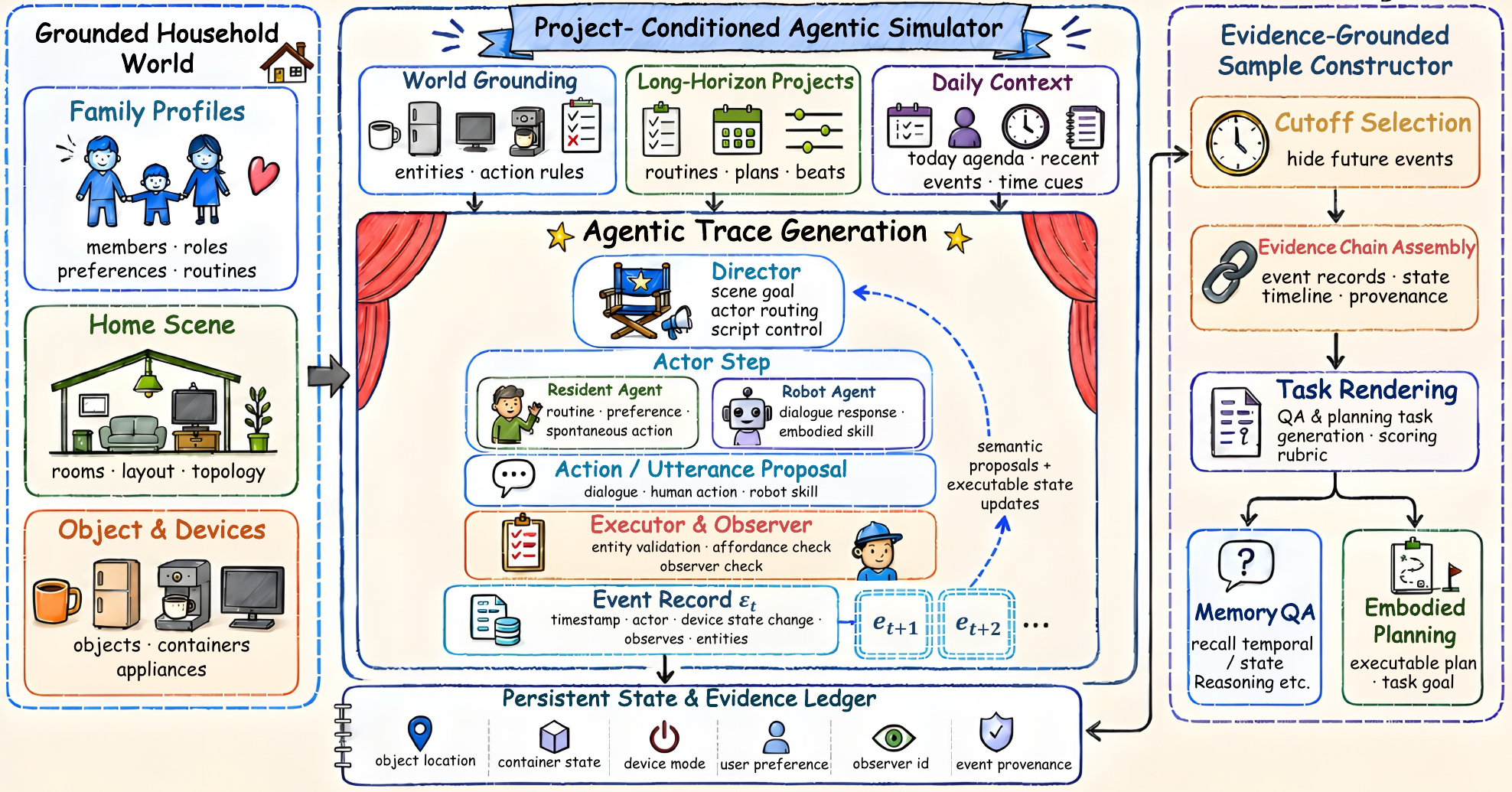}
    \caption{
    \textbf{Overview of the WorldLines construction framework.}
WorldLines builds long-horizon embodied traces from grounded household worlds, project-driven activities, and closed-loop state-changing interactions. These histories are converted into cutoff-controlled, evidence-linked samples for \textit{Memory QA} and \textit{Embodied Task Planning}, testing persistent world-state maintenance under partial observability.
    }
    \label{fig:framework}
\end{figure*}

\subsection{Benchmark Formulation}
\label{sec:benchmark-formulation}

\begin{wrapfigure}{r}{0.49\textwidth}
\vspace{-8pt}
\begin{casebox}[WorldLines Sample Snapshot]
\scriptsize
\texttt{May 1, 20:10}\quad Bob sets the weekend coffee routine to \texttt{08:30}.\\
\texttt{May 3, 23:42}\quad Bob mistakenly sets the kitchen coffee timer to \texttt{04:00}.\\
\texttt{May 4, 06:55}\quad The robot corrects the timer to Bob's weekday routine, \texttt{07:00}.\\[2pt]
\texttt{Cutoff: May 4, 07:05}\quad The user asks after the correction.\\
\texttt{Query}\quad \emph{``What time did Bob mistakenly set the coffee timer to, and what time did the robot correct it to?''}\\
\texttt{Evidence}\quad May 3 timer update $\rightarrow$ May 4 robot correction.\\
\texttt{Answer}\quad \texttt{04:00} and \texttt{07:00}.
\end{casebox}
\vspace{-10pt}
\caption{
\textbf{Example WorldLines sample.}
A timestamped query with its evidence chain and answer.
}
\label{fig:sample-snapshot}
\vspace{-8pt}
\end{wrapfigure}

WorldLines evaluates whether embodied agents can maintain household world states over long-term interactions. Each sample is derived from a multi-day household trace containing dialogue, human activity, robot actions, execution feedback, and object or device state changes. Beyond asking what an agent remembers, WorldLines tests whether the agent can use pre-cutoff visible history for question answering and state-aware planning.
Formally, each instance is represented as
\begin{equation}
\label{eq:worldlines-instance}
x_i = (\mathcal{H}_{<c_i}, q_i, S_{c_i}, \mathcal{E}_i, y_i^\star, \tau_i),
\end{equation}
where $\mathcal{H}_{<c_i}$ is the visible history before cutoff $c_i$, $q_i$ is the question or task instruction, $S_{c_i}$ is the ground-truth world state, $\mathcal{E}_i$ is the supporting evidence chain, $y_i^\star$ is the reference answer or plan, and $\tau_i$ denotes the task type. The agent observes only $(\mathcal{H}_{<c_i}, q_i)$; states and evidence are used only for evaluation.

Figure~\ref{fig:sample-snapshot} shows the structure of a concrete Memory QA sample. WorldLines contains two task families (Figure~\ref{fig:task_examples}). \textit{Memory QA} evaluates recovery of historical events, state changes, preferences, and routines. \textit{Embodied Task Planning} evaluates whether an agent can generate a state-consistent plan or next-step decision from visible history.

\begin{figure*}[t]
    \centering
    \includegraphics[width=0.96\textwidth]{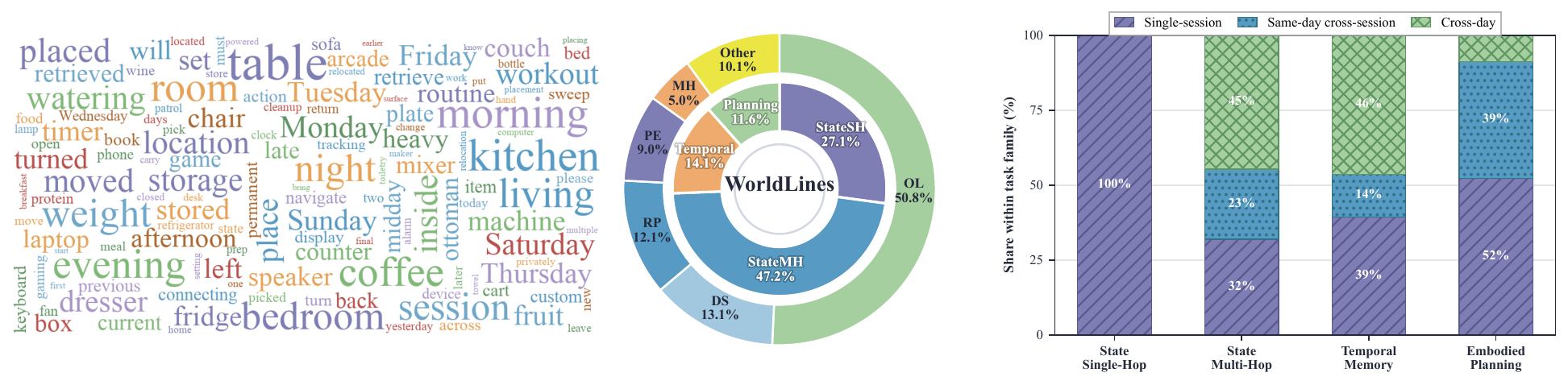}
    \caption{
    \textbf{Overview of WorldLines's statistical distributions.}
    \textbf{(Left)} Word distribution of benchmark queries and evidence text;
    \textbf{(Middle)} data distribution across task families and memory targets;
    and \textbf{(Right)} temporal-scope distribution across task families.
    }
    \label{fig:worldlines_benchmark_overview}
\end{figure*}

\subsection{Project-Driven Trace Generation}
\label{sec:trace-generation}

\noindent\textbf{World grounding.}
Figure~\ref{fig:framework} summarizes the WorldLines construction pipeline. World grounding defines what exists in each household and what can be executed. WorldLines uses Habitat/HSSD household scenes as grounded environments and curates interactive objects, receptacles, and controllable devices. These assets are converted into a semantic view for project generation and an executable scene view for validation. The executable view organizes object instances, device states, openable components, and action constraints in a scene-graph-style representation, allowing generated actions to be affordance-checked and written into replayable state trajectories.

\noindent\textbf{Project planning.}
Project planning gives each trace a persistent household thread. Instead of generating isolated daily events, WorldLines first creates long-term projects from the semantic view. Each project describes a multi-day life process, such as routine support, meal preparation, home organization, or device coordination, and specifies participants, relevant spaces and entities, temporal preferences, and constraints.

\noindent\textbf{Closed-loop trace simulation.}
Closed-loop simulation turns long-term projects into cross-day event streams that change the world state. For day $d$, the generation context is
\begin{equation}
\label{eq:daily-generation-context}
C_d = (\mathcal{W}, \mathcal{P}_d, S_{d-1}, N_{<d}),
\end{equation}
where $\mathcal{W}$ is the grounded world, $\mathcal{P}_d$ denotes active projects, $S_{d-1}$ is the accumulated state, and $N_{<d}$ contains carry-forward notes. Conditioned on $C_d$, the generator proposes dialogue, human activity, robot actions, and device operations. A deterministic executor checks entity references, affordances, and preconditions, and only valid outcomes are written as structured state changes.

\noindent\textbf{Carry-forward memory.}
Carry-forward memory makes traces continuous across days. After each day, WorldLines extracts notes from events, execution feedback, and state changes, including changed object or device states, user preferences, unresolved plans, recent events, and potential conflicts. These notes are passed to later days with the accumulated state, so future interactions depend on earlier household history.

\subsection{Evidence-Linked Task Construction}
\label{sec:task-construction}

After generating cross-day traces, WorldLines constructs evaluation samples from events and state changes that actually occurred. It indexes events, state timelines, entity histories, and action histories, then mines candidate questions and planning tasks. The LLM only rewrites structured candidates into natural language; the cutoff, ground truth, and evidence chain are determined programmatically.

Each sample has a context cutoff. The evaluated agent observes only pre-cutoff history, and the reference answer or plan must be supported by pre-cutoff evidence. This prevents future information leakage while enabling event-level retrieval and state-consistency evaluation.

Memory QA covers current states, overwritten states, multi-hop temporal reasoning, preferences, routines, and source-aware questions. Embodied Task Planning requires a plan or next-step decision from the current instruction, historical state, and executable constraints. Together, these tasks test whether long-horizon state maintenance supports reliable QA and embodied decision making.

\begin{figure*}[t]
    \centering
    \includegraphics[width=\textwidth]{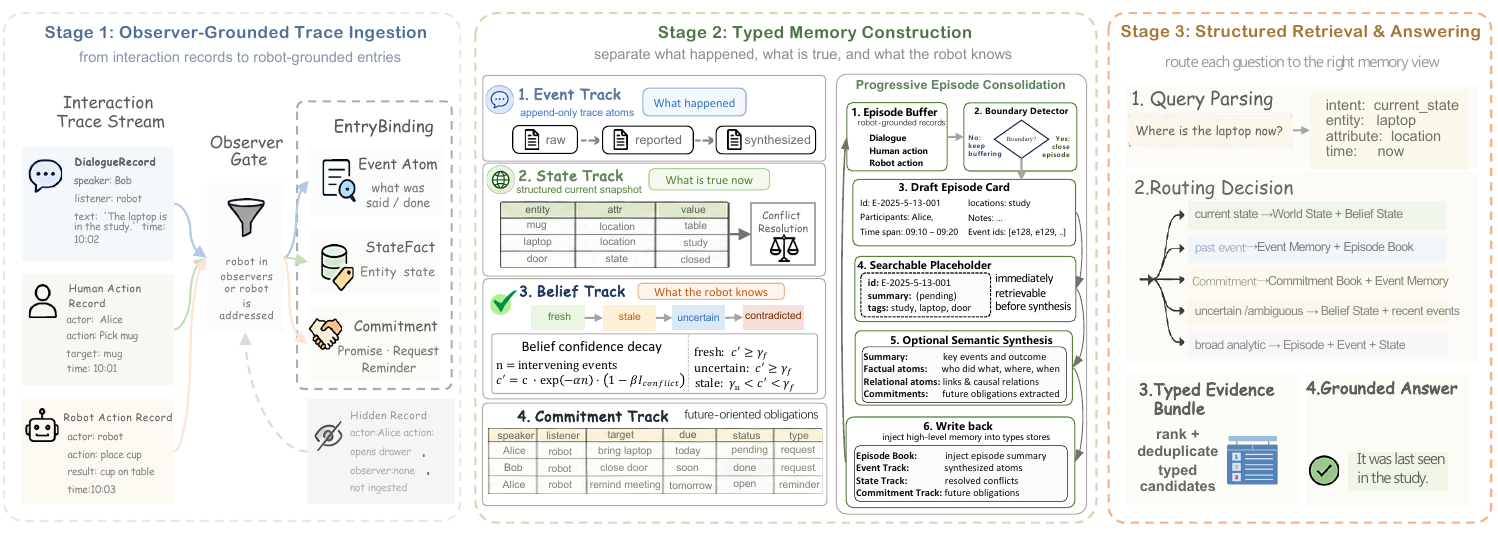}
    \caption{
    \textbf{Overview of ObsMem.}
    ObsMem organizes observer-grounded traces into typed views of events, states, beliefs, and commitments, then consolidates episodes for structured retrieval and grounded answering.
    }
    \label{fig:obsmem_framework}
\end{figure*}

\section{ObsMem: Observer-Grounded Memory}
\label{sec:obsmem}

ObsMem is motivated by the observation that long-horizon embodied memory is not simply about storing more history. A household trace contains different kinds of information: events that happened, world states that persist and change, the robot's epistemic confidence under partial observability, and future commitments that remain actionable. If all of these are compressed into a single text memory, the system cannot reliably tell whether a record is directly observed evidence, a reported claim, an overwritten state, or a future constraint.

ObsMem therefore treats memory as an online process that turns an interaction stream into typed evidence. Each new event is first gated by observation provenance, then updates different memory views according to its semantics. At query time, ObsMem does not perform one undifferentiated search over all text; instead, it composes the views needed by the task. We describe ObsMem along the lifecycle of a memory: how it is written, how it is updated, and how it is retrieved for answering and planning.

\subsection{Observer-Grounded Memory Ingestion}
\label{sec:obsmem-ingestion}

The first issue in embodied memory is observability. The robot may directly observe an object being moved, or it may only hear a user report where the object is. Both records may be useful later, but they should not carry the same reliability. ObsMem therefore uses an observer gate before writing events into memory, deciding whether an event is visible to the robot and whether it should be stored with observed or reported provenance.

Let $r$ denote the robot and $V_t$ the observer set of event $o_t$. An event enters the robot's memory only when $r\in V_t$. If the event is an utterance from a non-robot actor, ObsMem additionally creates a reported atom, explicitly separating what the robot heard someone say from what the robot directly observed:
\begin{equation}
\label{eq:observer-gate}
\Phi(o_t)=
\begin{cases}
\{e_t^{\mathrm{obs}}\}\cup
\mathbf{1}[\mathrm{utt}(o_t)\land \mathrm{actor}(o_t)\ne r]\{e_t^{\mathrm{rep}}\}, & r\in V_t,\\
\varnothing, & r\notin V_t.
\end{cases}
\end{equation}

After passing the gate, an event is written into the memory views that match its semantics. Every visible event enters the Event Track as traceable historical evidence. Executable actions additionally produce structured state facts in the State Track. Utterances that express requests, reminders, promises, or schedules are written into the Commitment Track. Thus, ObsMem preserves the semantic role of each record at write time, instead of asking retrieval to infer it later from plain text.

For example, if the robot observes itself placing a laptop on the sofa, the event creates both observed event evidence and a state fact such as \texttt{laptop.location=sofa}. If Bob merely says that the laptop is on the sofa, the utterance creates reported evidence, but it is not treated as an equally reliable direct observation. This distinction is preserved during later belief updates and query-time retrieval.

\subsection{Typed Memory Update}
\label{sec:obsmem-update}

After records are written, ObsMem updates them according to their roles. The key is not to maintain more memory slots, but to apply the right update rule to each type of information. The Event Track is append-only because historical evidence should not be overwritten. The State Track maintains both a current snapshot and a history because world states change over time. The Belief Track maintains epistemic reliability because the latest state is not always something the robot can confidently know. The Commitment Track keeps future constraints available for later QA and planning.

For the State Track, ObsMem represents each state-changing observation as a structured fact, where $i$ is the entity, $a$ is the attribute, $v_t$ is the new value, $\rho_t$ is the provenance, and $\tau_t$ is the timestamp. For each $(i,a)$, ObsMem maintains both a history $H_t(i,a)$ and a current snapshot $\hat{S}_t(i,a)$:
\begin{equation}
\label{eq:state-update}
f_t=(i,a,v_t,\rho_t,\tau_t),\qquad
H_t(i,a)=H_{t-1}(i,a)\cup\{f_t\},\qquad
\hat{S}_t(i,a)=\arg\max_{f\in H_t(i,a)} \tau(f).
\end{equation}

This design supports two kinds of questions. For a current-state query such as ``where is the laptop now?'', the system can read the current snapshot directly. For an overwritten-state query such as ``where was the laptop before Bob moved it?'', the system can still recover the historical trail. When a new fact conflicts with the current snapshot, the old fact is not deleted; it remains as overwritten evidence for explaining state transitions.

The Belief Track handles partial observability. It does not duplicate the world state; instead, it records the robot's epistemic status for each tracked fact. If a state was directly observed and no relevant intervention has occurred since, the belief is fresh. If the state was only reported, or if later events could have changed the entity outside the robot's observation, the belief becomes stale or uncertain. If contradictory evidence appears, the belief becomes contradicted.

Let $I_t$ be the intervening events that may affect $(i,a)$ since the last confirmation, $A_t$ the intervening actors, and $C_t$ the contradicting evidence. ObsMem updates the epistemic state with deterministic rules, where the thresholds are fixed implementation hyperparameters:
\begin{equation}
\label{eq:belief-transition}
z_t(i,a)=
\begin{cases}
\mathrm{contradicted}, & |C_t|>0,\\
\mathrm{uncertain}, & |I_t|\ge \lambda_I \ \lor\ (|I_t|\ge \lambda_m \land |A_t|\ge \lambda_A),\\
\mathrm{stale}, & |I_t|>0 \ \lor\ \rho_t\ne \mathrm{observed},\\
\mathrm{fresh}, & \text{otherwise}.
\end{cases}
\end{equation}

Continuing the laptop example, when Bob reports that the laptop is on the sofa, ObsMem can retain the reported evidence while marking the belief as less reliable than a direct observation. If the robot later observes the laptop on the table, the State Track updates the current location while the Event Track still preserves Bob's earlier report. If Bob later enters the room outside the robot's view, the Belief Track can mark \texttt{laptop.location} as uncertain, signaling to the answerer that the current state may have changed.

Finally, ObsMem performs episode-level consolidation to reduce fragmentation in low-level events. When an episode boundary is detected, the system creates an immediately retrievable episode card and optionally synthesizes summaries, factual atoms, relational atoms, and commitments back into the corresponding views. Importantly, summaries augment retrieval but do not replace the original events or state trails.

\subsection{Query-Time Retrieval and Answering}
\label{sec:obsmem-retrieval}

At query time, ObsMem composes evidence according to the question, rather than running a single similarity search over all memories. Different questions require different views: current-state queries need State and Belief, past-event queries need Event and Episode, commitment queries need Commitment and Event, and planning often requires current state, historical causes, future obligations, and uncertainty together.

Given a question or task instruction $q$, ObsMem first produces a query plan $p_q$ that identifies the intent, target entities, state attributes, temporal filters, and evidence views to access. Each view then performs its own retrieval. The State view first uses structured snapshot or point-in-time lookup before falling back to embedding search, while other views use their modality-specific indexes and filters.

Candidate evidence is the deduplicated union of the retrieval results from selected views. Here, $V(p_q)$ denotes the memory views selected by the query plan, and $R_v$ denotes the view-specific retriever:
\begin{equation}
\label{eq:routed-retrieval}
\mathcal{C}(q)=
\operatorname{dedup}\!\left(
\bigcup_{v\in V(p_q)}
R_v(q,p_q)
\right).
\end{equation}

An evidence selector then chooses a compact typed evidence bundle from the candidate set:
\begin{equation}
\label{eq:evidence-selection}
\hat{\mathcal{C}}_k=
\operatorname{Select}_{\theta}(q,p_q,\mathcal{C}(q),k).
\end{equation}
The selector does not merely keep the most semantically similar text. It favors complementary evidence across timestamps, entities, and memory views, so the answerer can jointly consider state, belief, and historical support.

For example, for ``Where is the laptop now?'', ObsMem routes the query as a current-state/location request, reads the laptop's State Track, and checks the Belief Track to determine whether the current location is reliable. For ``Who said it was on the sofa?'', the system turns to reported events. For an embodied planning request such as ``Please prepare the living room for movie night'', it combines current object states, relevant historical preferences, future commitments, and action preconditions to produce a more executable plan.

Thus, ObsMem's advantage is not simply storing more content, but preserving semantic structure throughout writing, updating, and retrieval. It distinguishes observed from reported, current from historical, known from uncertain, and remembered facts from executable constraints, enabling both evidence-grounded QA and state-aware embodied planning.

\section{Experiment}
\label{sec:experiment}

\subsection{Experiment Setup}
\label{sec:experiment-setup}

\begin{table*}[t]
\centering
\footnotesize
\setlength{\tabcolsep}{2pt}
\begin{tabular}{lcccccccc}
\toprule
\textbf{Method} & \textbf{Judge $\uparrow$} & \textbf{Perfect $\uparrow$} & \textbf{Sess. Any@5 $\uparrow$} & \textbf{Event R@5 $\uparrow$} & \textbf{StateMH-J $\uparrow$} & \textbf{StateMH-E $\uparrow$} & \textbf{StateSH-J $\uparrow$} & \textbf{Temp-J $\uparrow$} \\
\midrule
A-mem & 0.575 & 53\% & 0.839 & 0.355 & 0.540 & 0.216 & 0.550 & \textbf{0.692} \\
Mem0 & 0.554 & 53\% & 0.823 & 0.378 & 0.598 & 0.264 & 0.550 & 0.462 \\
GraphMem & 0.457 & 39\% & 0.806 & 0.243 & 0.529 & 0.184 & 0.417 & 0.359 \\
MemoryOS & 0.312 & 29\% & 0.452 & 0.085 & 0.287 & 0.086 & 0.350 & 0.308 \\
\textbf{ObsMem} & \textbf{0.713} & \textbf{69\%} & \textbf{0.879} & \textbf{0.537} & \textbf{0.762} & \textbf{0.452} & \textbf{0.667} & \textbf{0.667} \\
\bottomrule
\end{tabular}
\caption{
\textbf{Memory QA performance on WorldLines.}
All methods are evaluated on 310 Memory QA samples. We report overall QA quality, session/event retrieval, and state- or temporal-reasoning diagnostics. StateMH-E highlights event-level recall in the most evidence-demanding multi-hop state setting. Full family-level breakdowns are in Appendix~\ref{app:additional-metrics}.
}
\label{tab:overall-memory-qa}
\end{table*}

\begin{table*}[t]
\centering
\footnotesize
\setlength{\tabcolsep}{2pt}
\begin{tabular}{lccccccc}
\toprule
\textbf{Variant} & \textbf{Ablated Component} & \textbf{Judge $\uparrow$} & \textbf{$\Delta$ Judge} & \textbf{Perfect $\uparrow$} & \textbf{Event R@5 $\uparrow$} & \textbf{Hidden Judge $\uparrow$} & \textbf{Latency $\downarrow$} \\
\midrule
w/ Full ObsMem & -- & \textbf{0.699} & -- & \textbf{66\%} & \textbf{0.563} & \textbf{0.278} & 8.82 \\
w/o Belief & Belief-view retrieval & 0.651 & -0.048 & 63\% & 0.558 & 0.000 & 8.63 \\
w/o State & World-state retrieval & 0.597 & -0.102 & 56\% & 0.532 & 0.111 & 8.95 \\
w/o Consol. & Episode consolidation & 0.554 & -0.145 & 53\% & 0.419 & 0.167 & 9.16 \\
w/o Selector & Evidence selector & 0.435 & -0.264 & 40\% & 0.466 & 0.000 & \textbf{5.54} \\
\bottomrule
\end{tabular}
\caption{
\textbf{ObsMem ablation results on a 62-sample diagnostic QA subset.}
Each variant ablates one query-time memory view or mechanism while keeping the rest of the pipeline unchanged. Hidden Judge is computed on 6 hidden-until-observed questions. Latency is measured in seconds. Additional details are provided in Appendix~\ref{app:ablation-details}.
}
\label{tab:ablation}
\end{table*}

Figure~\ref{fig:worldlines_benchmark_overview} summarizes the scale and temporal coverage of WorldLines. Beyond raw trace size, WorldLines emphasizes delayed memory use and cross-day evidence, requiring systems to recover precise state-changing events rather than only broadly relevant sessions.

We evaluate Mem0~\cite{chhikara2025mem0}, A-mem~\cite{xu2025amem}, MemoryOS~\cite{kang2025memoryos}, GraphMem~\cite{hu2026doesmemoryneedgraphs}, and our proposed ObsMem on two evidence-linked tasks: \textit{Memory QA} and \textit{Embodied Task Planning}. All systems receive the same cutoff-controlled visible history. For QA, each system passes at most five retrieved records to the answer generator, and retrieval metrics are computed over these top-five records. Questions and reference answers are generated with GPT-4o-mini under evidence-linked constraints and manually verified against annotated supporting evidence; see Appendix~\ref{app:qa-verification}. All memory systems use \texttt{google/gemini-3.5-flash} for answer generation, and GPT-4o serves as the independent judge. Additional baseline and evaluation details are provided in Appendix~\ref{app:baseline-details}.

\noindent\textbf{Metrics.}
We report \textbf{Judge}, the normalized LLM-as-a-judge answer-quality score; \textbf{Perfect Rate}, the fraction of answers with a normalized score of 1.0; \textbf{Session Any@5}, the fraction of queries with at least one gold supporting session among the top-five retrieved records; and \textbf{Event R@5}, the average fraction of gold evidence events covered by the top-five records. Session-level recall measures coarse context coverage, while event-level recall evaluates precise embodied evidence recovery. Planning metrics are defined in Appendix~\ref{app:planning}. Efficiency and token-cost statistics are reported as supplementary analysis in Appendix~\ref{app:efficiency}.

\noindent\textbf{Judge reliability.}
To validate the automatic judge, we sample 80 system outputs covering all question families and evaluated methods. Human labels show substantial agreement (Fleiss' $\kappa=0.71$), and GPT-4o reaches 87.5\% agreement with the majority human label and 0.82 Spearman correlation with averaged human scores. Appendix~\ref{app:judge-validation} provides the full annotation protocol.

\subsection{Overall Evaluation}
\label{sec:overall-evaluation}

Table~\ref{tab:overall-memory-qa} shows that ObsMem achieves the strongest overall Memory QA performance. It obtains the highest Judge score and Perfect Rate, indicating that observer-grounded memory improves downstream answer quality. Compared with the strongest baseline on each metric, ObsMem improves Judge by 0.138 over A-mem and Event R@5 by 0.159 over Mem0. This suggests that its typed state trails and event-grounded retrieval help recover the concrete household evidence needed for answering.

The results also reveal a key property of WorldLines: retrieving a broadly relevant session is not sufficient for embodied memory. A-mem, Mem0, and GraphMem obtain relatively high Session Any@5, showing that they often reach the correct coarse temporal context. However, their Event R@5 is substantially lower than ObsMem. In dynamic household environments, a session may contain many object movements, device operations, and dialogue reports. Correct answering therefore requires identifying the exact state-changing event, not only retrieving semantically related text.

The question-family columns further show that ObsMem performs best on StateMultiHop and StateSingleHop questions, which require tracking mutable household states through one or more event transitions. We include StateMultiHop Event R@5 in the main table because this family requires recovering multiple state-changing events rather than only locating a broadly relevant session. Figure~\ref{case_study} illustrates this failure mode: a flat-text memory retrieves the salient routine but misses the anomalous state update, while ObsMem preserves the full state trail. The StateMH-E column shows a particularly large gap on event grounding for multi-hop state questions, where ObsMem improves over the strongest baseline by 0.188. A-mem remains competitive on TemporalMemory questions, suggesting that text-centric memories can be effective when temporal cues are explicit. Detailed event-level breakdowns and radar visualizations are provided in Appendix~\ref{app:additional-metrics} and Appendix~\ref{app:radar}. 

\begin{figure*}[t!]
\begin{casebox}[Case Study]
\refstepcounter{figure}\label{case_study}
\small
\texttt{Day 1 evening}\quad Bob's weekend coffee routine is set to \texttt{08:30}.\\
\texttt{Day 3 night}\quad Exhausted after work, Bob mistakenly sets the kitchen coffee timer to \texttt{04:00}.\\
\texttt{Day 4 early morning}\quad The robot detects the abnormal timer and corrects it to Bob's weekday routine, \texttt{07:00}.\\
\texttt{Query}\quad \emph{``What time did Bob mistakenly set the coffee timer to, and what time did the robot correct it to?''}

\vspace{4pt}
\noindent
\begin{tabularx}{\linewidth}{@{}>{\raggedright\arraybackslash}X@{\hspace{10pt}}>{\raggedright\arraybackslash}X@{}}
\textcolor{red!65!black}{\large$\blacktriangle$}\;
\textcolor{red!65!black}{\textbf{Flat-text memory (Mem0)}} &
\textcolor{green!45!black}{\large$\checkmark$}\;
\textcolor{green!45!black}{\textbf{ObsMem}} \\[2pt]

Retrieves the salient weekend routine \texttt{08:30} and the later correction to \texttt{07:00}. &
Maintains a typed state trail for the coffee timer: routine, anomalous update, and corrective update. \\[3pt]

\textbf{Answer:} \emph{``Bob set it to 08:30, and the robot corrected it to 07:00.''} &
\textbf{Answer:} \emph{``Bob mistakenly set it to 04:00, and the robot corrected it to 07:00.''} \\[3pt]

\textbf{Failure:} confuses a recurring routine with a one-off erroneous state. &
\textbf{Why:} separates \texttt{routine=08:30}, \texttt{anomaly=04:00}, and \texttt{correction=07:00} by entity, time, and actor. \\
\end{tabularx}
\end{casebox}
\end{figure*}

\subsection{Ablation Study}
\label{sec:ablation}

We conduct ablations on a 62-sample diagnostic subset covering the three question families and the main ObsMem mechanisms. The full ObsMem row is re-evaluated on this subset, so the results are not directly comparable to Table~\ref{tab:overall-memory-qa}. Table~\ref{tab:ablation} ablates four components: belief-view retrieval, world-state retrieval, episode consolidation, and LLM-based evidence selection. The w/o Belief and w/o State variants remove only query-time access to the corresponding memory views while keeping ingestion unchanged.

The evidence selector is the most critical component: removing it drops Judge from 0.699 to 0.435, despite moderate Event R@5, showing that answer quality depends on selecting and combining evidence across memory views, not only retrieval. Disabling episode consolidation causes the second largest drop, reducing Judge to 0.554 and Event R@5 to 0.419, indicating the value of episode synthesis and conflict resolution. Removing world-state retrieval hurts current-state reasoning, since object locations and device states must be reconstructed from raw events. The w/o Belief variant has a smaller overall impact, but reduces Hidden Judge from 0.278 to 0.000, suggesting that epistemic belief tracking is especially useful under partial observability.

\subsection{Downstream Embodied Planning Evaluation}
\label{sec:planning-results}

\begin{table}[t]
\centering
\small
\setlength{\tabcolsep}{3pt}
\begin{tabular}{lcccc}
\toprule
\textbf{Method} & \textbf{Plan $\uparrow$} & \textbf{State $\uparrow$} & \textbf{Precond. $\uparrow$} & \textbf{Mem. $\uparrow$} \\
\midrule
A-mem & 0.542 & 0.566 & 0.581 & 0.524 \\
Mem0 & 0.526 & 0.551 & 0.563 & 0.512 \\
GraphMem & 0.481 & 0.493 & 0.507 & 0.462 \\
MemoryOS & 0.337 & 0.361 & 0.376 & 0.319 \\
\textbf{ObsMem} & \textbf{0.684} & \textbf{0.721} & \textbf{0.702} & \textbf{0.690} \\
\bottomrule
\end{tabular}
\caption{
\textbf{Downstream embodied planning results.}
Full planning dimensions are provided in Appendix~\ref{app:planning}.
}
\label{tab:planning-compact}
\end{table}

Beyond Memory QA, we evaluate whether retrieved long-term memory can support executable, state-aware household planning. This setting is more demanding than QA because an agent must not only recall relevant evidence, but also use it to check object locations, container states, device states, and action preconditions before producing an action sequence. Although the planning set contains 21 samples, each instance is action-dense, with an average of 7.6 target actions and 3.1 remembered state constraints.

Table~\ref{tab:planning-compact} reports planning results. ObsMem obtains the highest Plan Judge score and performs especially well on state consistency, precondition validity, and memory use. These gains suggest that typed state trails help translate remembered household states into executable plans. In contrast, text-centric or graph-expanded memory systems can retrieve useful task context but are less reliable at converting retrieved information into explicit state constraints. Full planning dimensions are provided in Appendix~\ref{app:planning}.




\section{Conclusion}
We introduced \textbf{WorldLines}, a benchmark for evaluating long-horizon stateful embodied agents in dynamic and partially observable household environments. Unlike prior memory or embodied-task benchmarks, WorldLines focuses on whether agents can maintain persistent world states across dialogue, human activity, robot actions, device changes, and execution feedback, and use them for Memory QA and Embodied Task Planning. We further proposed \textbf{ObsMem}, an observer-grounded memory framework that separates event evidence, structured world states, and agent beliefs to support state-aware reasoning under partial observability. Experiments show that existing memory systems struggle with overwritten states, uncertainty, and translating long-term memory into embodied decisions, highlighting the need for memory architectures designed for stateful embodied interaction.

\section{Limitations}

WorldLines is constructed in simulated household environments and does not fully cover perception noise, actuation errors, or open-ended human behavior in real homes. This controlled setting enables precise annotation of evidence chains, cutoffs, state changes, and executable constraints for systematic evaluation. Future work can extend WorldLines to real robot logs, visual observations, and full physical simulation.

ObsMem is designed for structured embodied traces with entity identifiers, visibility annotations, and action schemas. In real deployment, these signals would need to be provided by perception, localization, and grounding modules. ObsMem also introduces additional latency from typed retrieval and belief-aware evidence selection, motivating more efficient retrieval and integration with visual perception and execution feedback.

\newpage
\bibliographystyle{assets/plainnat}
\bibliography{paper}

\clearpage
\beginappendix
\section{Additional Experimental Details}
\label{app:experiment-details}

\subsection{Baseline Implementation Details}
\label{app:baseline-details}

We evaluate Mem0, A-mem, MemoryOS, and GraphMem as representative long-term memory baselines. When an official implementation is available, we use the official codebase and retain the default memory-update procedure recommended by the original method. When a method requires a system-specific memory construction procedure, we keep that procedure unchanged because memory updating is part of the method being evaluated. All methods receive the same cutoff-controlled visible household history before each query.

\begin{table*}[t]
\centering
\small
\setlength{\tabcolsep}{4pt}
\begin{tabular}{lp{0.19\textwidth}p{0.31\textwidth}p{0.22\textwidth}}
\toprule
\textbf{Method} & \textbf{Implementation} & \textbf{Memory Update} & \textbf{Retrieval / Context Cap} \\
\midrule
Mem0 & Official & Default extraction/update pipeline & Top-5 final records \\
A-mem & Official & Agentic memory evolution & Top-5 final records \\
MemoryOS & Official & Hierarchical memory update & Top-5 final records \\
GraphMem & Reimplemented from specification & Graph construction and expansion & Top-5 graph-expanded records \\
ObsMem & Ours & Observer-grounded state-trail update & Up to 5 typed evidence records \\
\bottomrule
\end{tabular}
\caption{
\textbf{Baseline implementation and context control.}
All methods receive the same cutoff-controlled visible history. Retrieval can use method-specific internal mechanisms, but the final answer-generation context is capped to at most five records.
}
\label{tab:appendix-baseline-impl}
\end{table*}

For answer generation, we enforce a shared context budget across systems. Each method may retrieve using its own internal scoring or expansion mechanism, but only the top five final retrieved records are passed to the answer generator. Retrieval metrics are computed over the same top-five records. For GraphMem, graph expansion is allowed during retrieval, but the final graph-expanded context is capped to five retrieved records before answer generation. This ensures that downstream answer quality is not driven by unequal context length.

ObsMem differs from generic text-memory baselines by using observer-grounded storage, typed state trails, and belief-aware retrieval. Its retriever may return fewer than five records when fewer high-confidence typed evidence records are available.

\subsection{QA Verification Protocol}
\label{app:qa-verification}

Benchmark questions and reference answers are generated under evidence-linked constraints and then manually verified against annotated supporting evidence. During verification, annotators check that each QA item satisfies three criteria: (1) the question is answerable before the context cutoff, (2) the reference answer is entailed by annotated supporting evidence, and (3) answering the question does not require post-cutoff information or evaluator-only hidden state. Items failing any criterion are revised or removed.

\subsection{Judge Validation Protocol}
\label{app:judge-validation}

We validate the LLM-as-a-judge protocol on 80 sampled system outputs. The subset is selected to cover all evaluated methods and all question families. Five human annotators independently score each answer using the same 0--3 correctness rubric used by the automatic judge. We map scores into three categories: incorrect, partially correct, and correct, and compare GPT-4o judge decisions against the majority human label.

Human annotations show substantial agreement, with Fleiss' $\kappa=0.71$. GPT-4o achieves 87.5\% agreement with the majority human label and a Spearman correlation of 0.82 with averaged human scores. These results suggest that GPT-4o judge scores provide a reasonable proxy for answer correctness, while retrieval-based metrics such as Session Any@5 and Event R@5 provide judge-independent evidence-grounding diagnostics.

\begin{table}[t]
\centering
\footnotesize
\setlength{\tabcolsep}{2.5pt}
\begin{tabular}{cl}
\toprule
\textbf{Score} & \textbf{Meaning} \\
\midrule
0 & Incorrect or contradicts reference evidence \\
1 & Partially relevant but misses key evidence \\
2 & Mostly correct with minor omissions \\
3 & Fully correct and evidence-consistent \\
\bottomrule
\end{tabular}
\caption{
\textbf{Judge rubric for Memory QA answer correctness.}
The automatic judge and human annotators use the same 0--3 rubric.
}
\label{tab:appendix-judge-rubric}
\end{table}

\subsection{Radar Visualization by Embodied Memory Type}
\label{app:radar}

\begin{figure*}[t]
    \centering
    \includegraphics[width=0.92\textwidth]{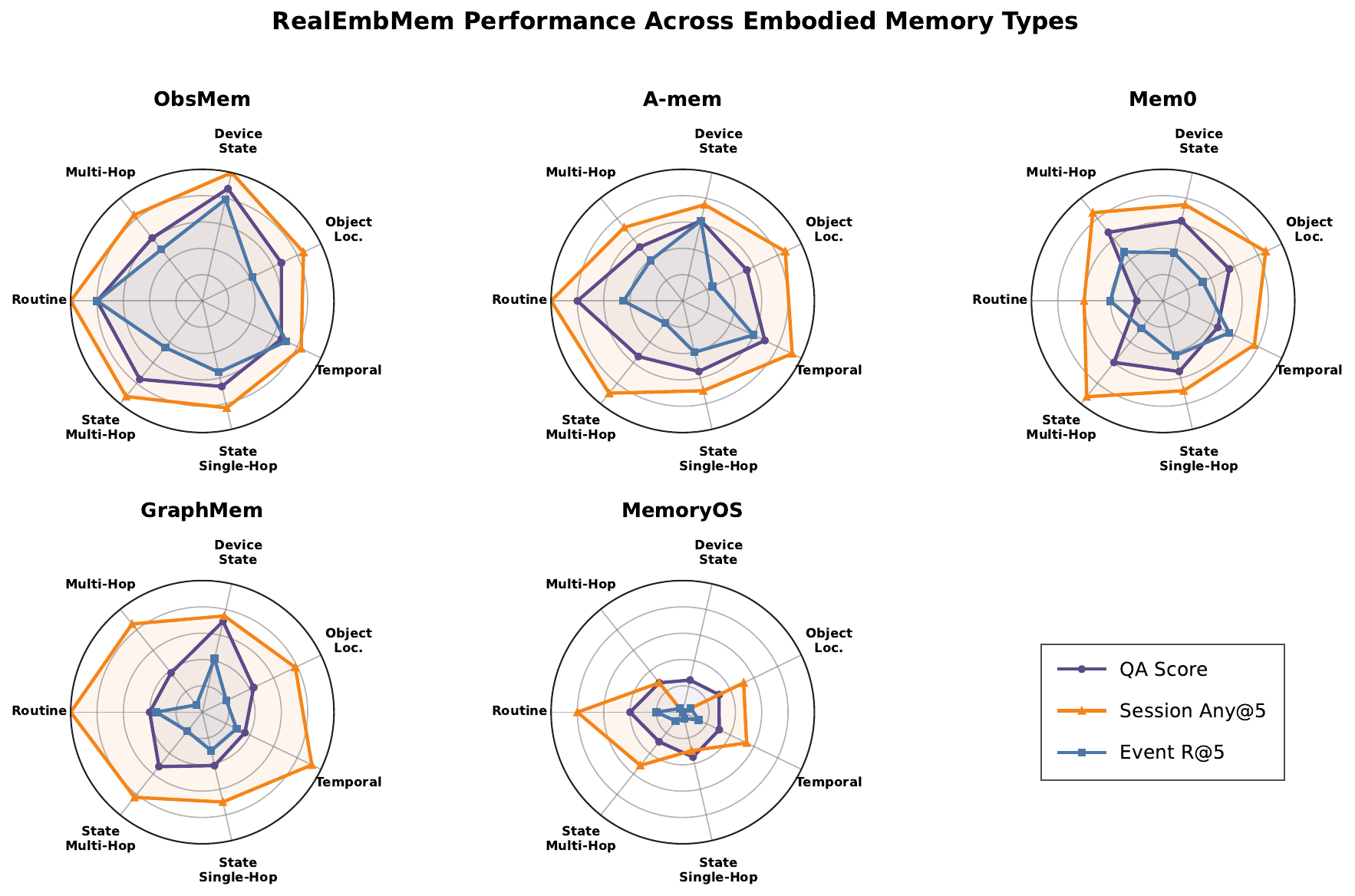}
    \caption{
    \textbf{Performance across embodied memory question types.}
    Each radar chart corresponds to one memory system. Axes denote representative embodied memory categories, and curves report QA score, session recall, and event recall. ObsMem shows a more balanced profile across state-centric categories, while generic memory systems often retrieve coarse sessions without matching the same level of event grounding.
    }
    \label{fig:appendix-question-type-radar}
\end{figure*}

Figure~\ref{fig:appendix-question-type-radar} provides a detailed visualization of method behavior across embodied memory types. The contrast between Session Any@5 and Event R@5 shows that methods can retrieve broadly relevant sessions without recovering the exact state-changing events required by the answer.

\subsection{Ablation Variant Details}
\label{app:ablation-details}

The ablation study removes one central ObsMem mechanism at a time while keeping the rest of the ingestion, retrieval, answer-generation, and judging pipeline unchanged. We run the ablation on a 62-sample diagnostic QA subset covering StateMultiHop, StateSingleHop, and TemporalMemory questions.

\begin{table}[t]
\centering
\small
\setlength{\tabcolsep}{6pt}
\begin{tabular}{lc}
\toprule
\textbf{Subset Type} & \textbf{N} \\
\midrule
StateMultiHop & 29 \\
StateSingleHop & 20 \\
TemporalMemory & 13 \\
\midrule
Hidden-until-observed & 6 \\
\bottomrule
\end{tabular}
\caption{
\textbf{Ablation diagnostic subset composition.}
Question-family counts sum to 62. Hidden-until-observed is a visibility subset used for the Hidden Judge diagnostic.
}
\label{tab:appendix-ablation-subset}
\end{table}

\begin{table*}[t]
\centering
\small
\setlength{\tabcolsep}{5pt}
\begin{tabular}{lp{0.24\textwidth}p{0.54\textwidth}}
\toprule
\textbf{Variant} & \textbf{Removed Component} & \textbf{Implementation Change} \\
\midrule
NoBelief & Belief retrieval & Belief records are still maintained during ingestion, but the router does not surface belief candidates to the answerer. \\
NoState & World-state retrieval & World-state facts are still maintained for diagnostics, but state candidates are removed from query-time retrieval. \\
NoConsol & Episode consolidation & Episode synthesis and state-conflict resolution are disabled; only minimal structural placeholders are retained for interface compatibility and contain no additional semantic summaries. \\
NoSelector & Evidence selector & The LLM reranker is bypassed, and candidates are passed in router-emitted order. \\
\bottomrule
\end{tabular}
\caption{
\textbf{ObsMem ablation variants.}
Each variant isolates one mechanism in the ObsMem memory pipeline without changing the benchmark samples, backbone model, judge, or answer-generation prompt format.
}
\label{tab:appendix-ablation-variants}
\end{table*}

\subsection{Downstream Embodied Planning Probe}
\label{app:planning}

We score plans along four dimensions. \textit{State consistency} checks whether the plan respects remembered object, container, and device states. \textit{Precondition validity} checks whether each proposed action is executable under the current state. \textit{Memory use} checks whether relevant long-term household context is incorporated. \textit{Action order} checks whether the proposed steps form a coherent executable sequence. Plan Judge is the average of these four scores.

\begin{table*}[t]
\centering
\footnotesize
\setlength{\tabcolsep}{2.5pt}
\resizebox{\textwidth}{!}{%
\begin{tabular}{lcccccc}
\toprule
\textbf{Method} & \textbf{N} & \textbf{Plan Judge $\uparrow$} & \textbf{State Consistency $\uparrow$} & \textbf{Precondition Validity $\uparrow$} & \textbf{Memory Use $\uparrow$} & \textbf{Action Order $\uparrow$} \\
\midrule
ObsMem & 21 & \textbf{0.684} & \textbf{0.721} & \textbf{0.702} & \textbf{0.690} & \textbf{0.676} \\
A-mem & 21 & 0.542 & 0.566 & 0.581 & 0.524 & 0.553 \\
Mem0 & 21 & 0.526 & 0.551 & 0.563 & 0.512 & 0.535 \\
GraphMem & 21 & 0.481 & 0.493 & 0.507 & 0.462 & 0.496 \\
MemoryOS & 21 & 0.337 & 0.361 & 0.376 & 0.319 & 0.352 \\
\bottomrule
\end{tabular}
}
\caption{
\textbf{Downstream planning probe on 21 action-dense samples.}
Planning requires agents to use remembered household states for executable, state-aware action decisions.
}
\label{tab:appendix-planning}
\end{table*}

The planning results serve as a smaller-sample downstream planning probe. Each planning instance contains an average of 7.6 target actions and requires checking 3.1 remembered state constraints on average, including object locations, container states, device settings, and action preconditions. On this action-dense set, ObsMem shows promising gains across all dimensions, especially state consistency and precondition validity. This suggests that structured state trails may help generate more state-aware plans.

\subsection{Full Efficiency and Context-Cost Statistics}
\label{app:efficiency}

\begin{table*}[t]
\centering
\small
\setlength{\tabcolsep}{5pt}
\begin{tabular}{lcccc}
\toprule
\textbf{Method} & \textbf{Avg. Latency (s) $\downarrow$} & \textbf{Prompt Tok. $\downarrow$} & \textbf{Completion Tok. $\downarrow$} & \textbf{Avg. Passed Records} \\
\midrule
A-mem & \textbf{3.35} & 625 & 498 & 5.0 \\
Mem0 & 4.57 & 513 & 539 & 5.0 \\
MemoryOS & 4.04 & \textbf{507} & 378 & 5.0 \\
GraphMem & 4.43 & 2057 & 388 & 5.0 \\
ObsMem & 5.13 & 562 & \textbf{319} & 3.4 \\
\bottomrule
\end{tabular}
\caption{
\textbf{Efficiency and context cost.}
All systems pass at most five retrieved records to the answer generator. Avg. Passed Records measures final context size rather than retrieval correctness; lower values indicate fewer records passed to the generator, not necessarily better retrieval. ObsMem often returns fewer records due to typed retrieval, while incurring higher latency from structured state and belief-aware retrieval.
}
\label{tab:appendix-efficiency}
\end{table*}

All systems use the same maximum answer-generation budget of five retrieved records. ObsMem often returns fewer records because its typed retriever abstains from adding weakly matched evidence. GraphMem uses the largest prompt context: although it is capped to five final retrieved records, each graph-expanded record may contain neighboring node descriptions, resulting in a larger prompt context.

\subsection{Additional Metrics}
\label{app:additional-metrics}

We report additional question-family event-level metrics in Table~\ref{tab:appendix-question-family-event}. These diagnostics complement the compact main-table results and further show the gap between coarse session retrieval and precise event-level grounding.

\begin{table*}[t]
\centering
\small
\setlength{\tabcolsep}{4pt}
\begin{tabular}{lcccccc}
\toprule
\textbf{Method} 
& \multicolumn{2}{c}{\textbf{StateMultiHop (145)}} 
& \multicolumn{2}{c}{\textbf{StateSingleHop (100)}} 
& \multicolumn{2}{c}{\textbf{TemporalMemory (65)}} \\
\cmidrule(lr){2-3}
\cmidrule(lr){4-5}
\cmidrule(lr){6-7}
& \textbf{Judge $\uparrow$} & \textbf{Event R@5 $\uparrow$}
& \textbf{Judge $\uparrow$} & \textbf{Event R@5 $\uparrow$}
& \textbf{Judge $\uparrow$} & \textbf{Event R@5 $\uparrow$} \\
\midrule
ObsMem & \textbf{0.762} & \textbf{0.452} & \textbf{0.667} & \textbf{0.556} & 0.667 & \textbf{0.708} \\
A-mem & 0.540 & 0.216 & 0.550 & 0.400 & \textbf{0.692} & 0.596 \\
Mem0 & 0.598 & 0.264 & 0.550 & 0.425 & 0.462 & 0.558 \\
GraphMem & 0.529 & 0.184 & 0.417 & 0.300 & 0.359 & 0.288 \\
MemoryOS & 0.287 & 0.086 & 0.350 & 0.050 & 0.308 & 0.135 \\
\bottomrule
\end{tabular}
\caption{
\textbf{Full question-family performance with event-level grounding.}
The main paper reports a compact subset of these family-level metrics; this appendix table provides the complete Event R@5 breakdown.
}
\label{tab:appendix-question-family-event}
\end{table*}

\section{Replay and Edited Scene Examples}
\label{app:replay-scene-examples}

\begin{figure*}[t]
\centering
\IfFileExists{figures/replay.pdf}{%
    \includegraphics[width=0.96\textwidth]{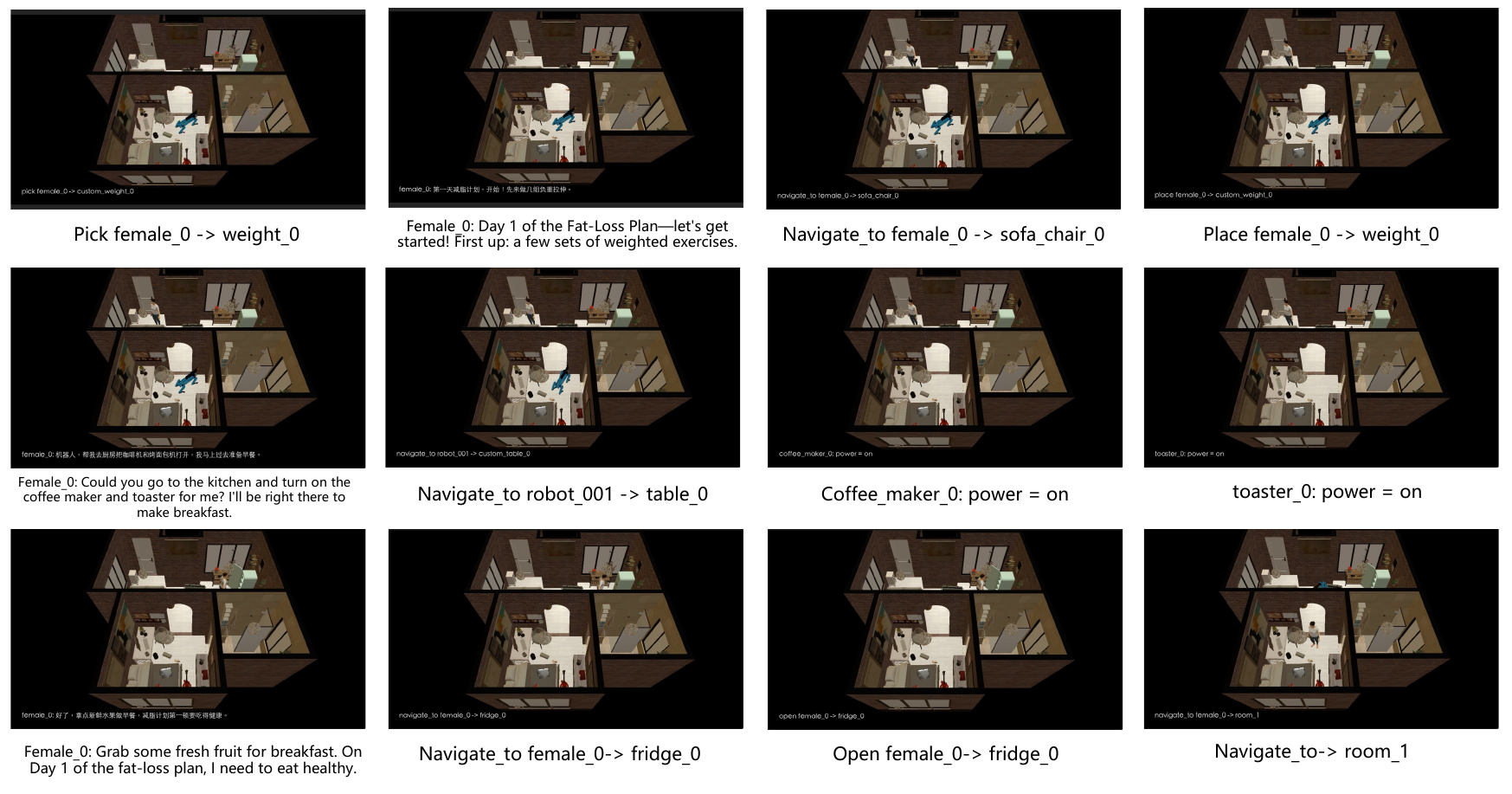}
}{%
    \fbox{\parbox[c][0.26\textheight][c]{0.88\textwidth}{\centering
    Placeholder for \texttt{figures/replay.pdf}\\
    Replay screenshots will be inserted here.}}
}
\caption{
\textbf{Replay examples.}
Generated household traces can be replayed in the corresponding edited
Habitat scene to visualize robot \& human npc actions, object relocation. 
}
\label{fig:appendix-replay-examples}
\end{figure*}

\begin{figure*}[t]
\centering
\IfFileExists{figures/scene_editing.png}{%
    \includegraphics[width=\textwidth]{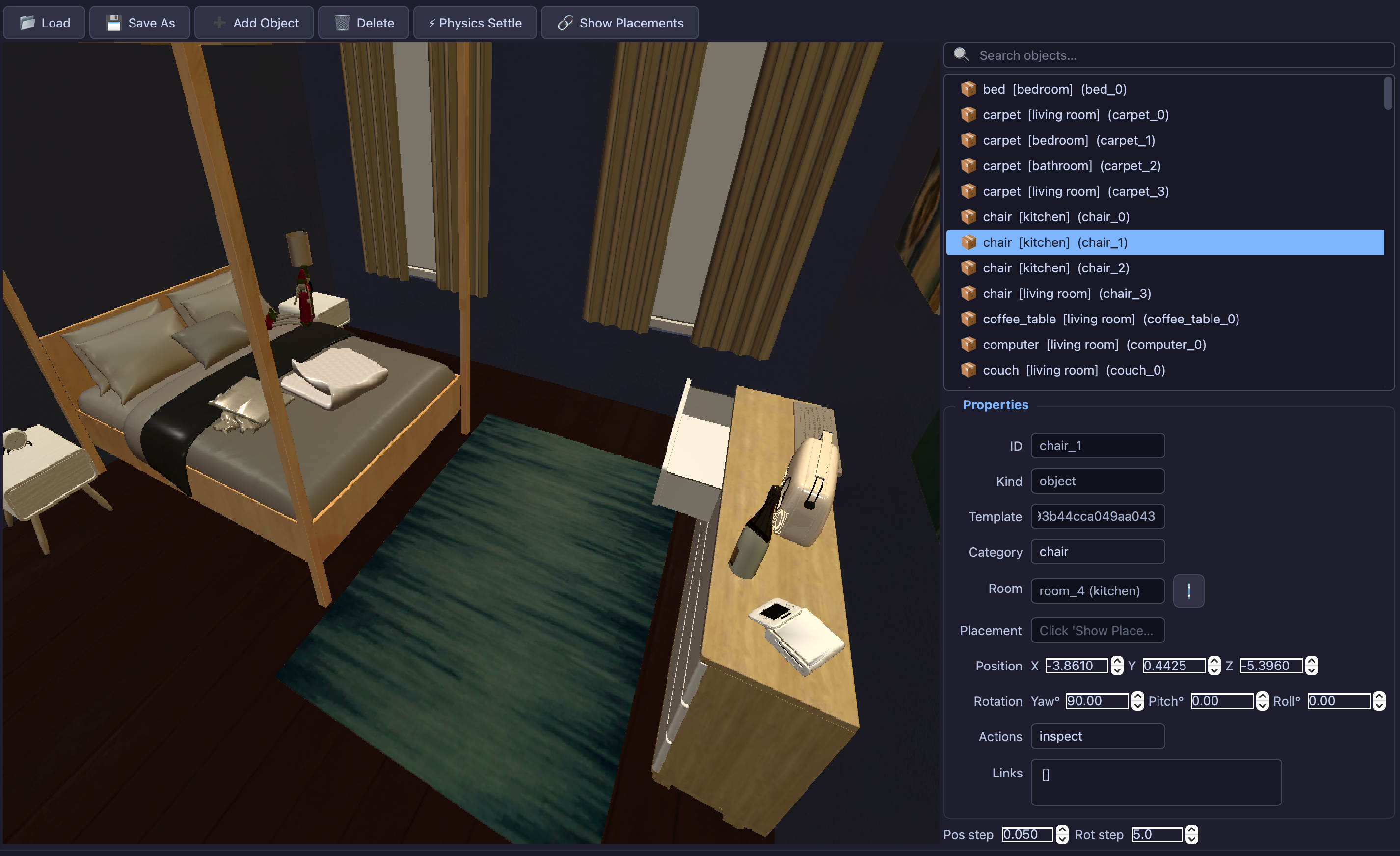}
}{%
    \fbox{\parbox[c][0.26\textheight][c]{0.88\textwidth}{\centering
    Placeholder for \texttt{figures/scene\_editing.pdf}\\
    Edited-scene interface screenshot will be inserted here.}}
}
\caption{
\textbf{Edited Habitat scene example.}
WorldLines manually curates Habitat/HSSD scenes by selecting household-relevant
objects, receptacles, controllable devices, and actor roles before trace
generation. This scene editing defines executable affordances; it does not
hand-author benchmark questions or reference answers.
}
\label{fig:appendix-scene-editing}
\end{figure*}

WorldLines is grounded in Habitat/HSSD household scenes rather than text-only
interaction logs. For each family, we manually curate the base scene by
selecting interaction-relevant objects, valid receptacles, controllable
devices, and household roles. The generated event traces can then be replayed
against the edited scene state. These visualizations are used for qualitative
inspection and illustration; evaluation itself uses the structured event logs,
state transitions, cutoff annotations, and evidence links described in the main
paper.
Figure~\ref{fig:appendix-replay-examples} provides replay examples, and
Figure~\ref{fig:appendix-scene-editing} shows a placeholder for the edited
scene interface screenshot.

\section{Benchmark Taxonomy and Action Space}
\label{sec:appendix-taxonomy}

WorldLines uses a controlled vocabulary to generate long-horizon household traces. It consists of three parts: project types, memory targets, and executable skills.

\noindent\textbf{Project Types.}
Project types define the high-level household themes used to organize long-horizon traces:
\begin{itemize}
    \item household\_routine: recurring activities that structure daily household life.
    \item routine\_support: support for an individual's repeated habits, schedules, or preparation needs.
    \item household\_organization: ongoing organization, storage, tidying, or rearrangement themes.
    \item health\_lifestyle: health, exercise, recovery, or lifestyle-related household projects.
    \item meal\_preparation: food, drink, grocery, meal planning, or kitchen-related routines.
    \item work\_study\_support: work, study, documents, workspace, or deadline-driven support.
    \item digital\_device\_coordination: smart-device, remote-control, or digital state coordination.
    \item family\_comfort: comfort, relaxation, entertainment, or family well-being activities.
\end{itemize}

\noindent\textbf{Memory Targets.}
Memory targets specify the intended memory challenges within each project:
\begin{itemize}
    \item object\_location: where movable objects are placed, moved, hidden, or retrieved.
    \item temporal\_state: time-sensitive commitments, deadlines, schedules, and recent changes.
    \item device\_state: functional states of controllable devices, such as power, mode, or timer.
    \item preference: user preferences revealed through dialogue, routines, or repeated choices.
    \item routine: repeated behavior patterns, habits, or ``usual'' household arrangements.
    \item planning\_dependency: facts that affect later planning, task ordering, or delayed execution.
    \item hidden\_state: state changes that are not directly observed by the robot or another actor.
    \item social\_context: social commitments, interpersonal context, or family coordination needs.
\end{itemize}

\noindent\textbf{Action Space.}
The action space defines the executable skill interface used by the closed-loop trace generator:
\begin{itemize}
    \item navigate\_to: move an actor to a room, furniture, device, or object anchor before physical interaction.
    \item inspect: observe a target object, receptacle, device, or furniture item to reveal state or contextual information.
    \item pick: pick up a movable object or device node, subject to co-location and empty-hand constraints.
    \item place: place the held object onto or into a valid destination surface, receptacle, furniture, or device.
    \item open: open an openable container, furniture item, or device component.
    \item close: close an openable container, furniture item, or device component.
    \item set\_device\_state: modify a supported smart-device state field, such as power, mode, temperature, or timer.
    \item handoff: transfer a held object from one co-located actor to another actor with empty hands.
\end{itemize}

\section{Prompt Templates}
\label{sec:appendix-prompts}

This appendix presents the key prompt templates used in our data generation pipeline and evaluation protocol.
Placeholders in \texttt{\{curly braces\}} are filled programmatically at runtime.

\subsection{Project Candidate Generation (Stage~2)}
\label{sec:prompt-stage2}

The following system prompt instructs the LLM to generate diverse
candidate household life-theme projects for a given family and home
environment. The LLM outputs semantic project descriptions without
binding to concrete object IDs or action sequences; downstream stages
ground each project into executable traces. Figure~\ref{fig:stage2-project-prompt}
shows the condensed prompt excerpt.

\begin{figure*}[t]
\begin{promptbox}[Stage~2: Project Candidate --- System Prompt]
\begin{lstlisting}[style=promptstyle]
## Role
You are a household life-theme ideation agent for an Embodied Memory Benchmark.
## Task
Your task is to generate diverse candidate long-running life themes for a given family and home. A project represents a LIFE THEME -- an ongoing background concern, seasonal goal, evolving habit, or household phase that shapes what naturally happens over many days. Think of a project as "what this family is going through lately", NOT as "what the robot should do every day".
## Good project framing
  - "Alice is doing a two-week morning fitness and healthy-breakfast push. The project has a clear goal, recurring morning pressure, and can naturally create memory opportunities around exercise setup, kitchen device states, and adjusted routines."
  - "Alice is reorganizing bedroom and bathroom storage to reduce morning friction. The project can create temporary item relocations, hidden or partially observed human actions, and later questions about where things were put."
## Bad project framing
  - "Every morning, prepare coffee and set the radio"
  - "Robot should pick up dishes and load dishwasher"
## Key idea
  - semantic theme, not entity binding Describe what the household is going through and what kinds of memory/action opportunities may arise. Do NOT choose concrete object IDs or robot skill families. Specific objects and executable actions will be selected later by the Stage 3 trace simulator.
## Selected rules
- Projects should be diverse in type, tempo, and scope.
- Projects should collectively explore: physical object movement, device state, dialogue-revealed preferences, hidden human actions, cross-day temporal dependencies, interruptions/recovery, and planning dependencies.
- Do not make a project sound like a scripted routine that repeats identically every day.
\end{lstlisting}
\vspace{1pt}
\end{promptbox}
\caption{Excerpt of the Stage~2 project candidate generation prompt.}
\label{fig:stage2-project-prompt}
\end{figure*}

\subsection{Project Day Beat Planning (Stage~3A)}
\label{sec:prompt-stage3a}

Stage~3A expands each selected household project into a sparse
multi-day semantic trajectory. It decides when a project is established,
interrupted, recovered, or completed, while deliberately avoiding
concrete sessions, rooms, actions, or state changes. Figure~\ref{fig:stage3a-day-beat-prompt}
shows the corresponding prompt excerpt.

\begin{figure*}[t]
\begin{promptbox}[Stage~3A: Project Day Beat --- System Prompt]
\begin{lstlisting}[style=promptstyle]
## Role
You are an embodied memory benchmark data generator.

## Task
Create a high-level, multi-day Project Day Beat plan for ONE household life-theme project. You decides only the semantic rhythm of a project across days. It must NOT decide concrete sessions, rooms, entity IDs, action sequences, or state diffs. Later components will ground each beat into sessions, executable actions, and validated state transitions.

## Output
Return a JSON array of selected project beat objects. Each object contains:
- day_index
- phase: establish, active_progress, obstacle, recovery, or completion
- milestone
- disruption
- memory_focus
- note

## Selected rules
- Choose day_index values within the requested horizon and keep them strictly increasing.
- Use temporal constraints to create natural gaps between meaningful project beats.
- If there is an obstacle, include a later recovery or adaptation day.
- Do not assume ideal project execution on every active day. Real household life may reshape the project through fatigue, stress, visitors, errands, forgotten preparation, or schedule conflicts.

## Allowed memory targets
object_location, temporal_state, device_state, preference, routine, planning_dependency, hidden_state, social_context
\end{lstlisting}
\end{promptbox}
\caption{Excerpt of the Stage~3A project day beat planning prompt.}
\label{fig:stage3a-day-beat-prompt}
\end{figure*}

\subsection{Session Intent Planning (Stage~3B)}
\label{sec:prompt-stage3b}

Stage~3B converts the project beats active on a given day into ordered household sessions. It plans life situations, timing, visibility, and narrative pressure, but still avoids executable actions and concrete object manipulation. Figure~\ref{fig:stage3b-session-intent-prompt} shows the prompt excerpt used at this stage.

\begin{figure*}[t]
\begin{promptbox}[Stage~3B: Session Intent --- System Prompt]
\begin{lstlisting}[style=promptstyle]
## Role
You are the day planner for a household long-term memory benchmark.

## Task
Arrange ONE day's project beats into ordered household sessions. You only plan life situations and narrative pressure. Do NOT write concrete actions, object IDs, dialogue, or state diffs; the closed-loop trace generator will ground each session later.

## Session constraints
- Every project beat must be covered by at least one session.
- session_type must be one of: morning, daytime, evening, late_night.
- Every session must include start_time and end_time inside the chosen time window.
- project_id must come from today's project beats. The same project may appear in multiple sessions.

## Narrative constraints
- purpose should describe why this life moment happens, what pressure or change shapes it, and how it relates to the project.
- narrative_hook must explain the concrete past fact, commitment, preference, object state, or time pressure that affects this session.
- Make the day life-like, not a clean project plan. Use realistic volatility: fatigue, meetings, weather, visitors, errands, missing objects, device problems, delays, preference changes, unfinished tasks, or misunderstandings.

## Visibility modes
- mixed: NPC and robot share the scene or coordinate by dialogue/remote instruction.
- npc_only_hidden: NPC acts privately; the robot does not know yet.
- robot_only: robot acts while NPC is away, asleep, or uninvolved.

## Output
Return a JSON object with a sessions array. Each session contains session_type, start_time, end_time, project_id, purpose, visibility_mode, and narrative_hook.
\end{lstlisting}
\end{promptbox}
\caption{Excerpt of the Stage~3B session intent planning prompt.}
\label{fig:stage3b-session-intent-prompt}
\end{figure*}

\subsection{Closed-Loop Agentic Trace Generation}
\label{sec:prompt-closed-loop-trace}

The final trace generator uses a closed-loop director--actor--executor
protocol rather than a single scriptwriter. The director first creates a
compact session setup, actors then propose executable event sequences, and
the deterministic executor applies state changes and derives visibility.
Figures~\ref{fig:director-setup-prompt}, \ref{fig:actor-turn-prompt},
and~\ref{fig:session-examiner-prompt} show concise prompt excerpts
for the three LLM-facing components.

\begin{figure*}[t]
\begin{promptbox}[Director Setup --- System/User Prompt Excerpt]
\begin{lstlisting}[style=promptstyle]
## Role
You are the Session Director for an embodied memory benchmark.
You are not an in-world actor. You control story pacing, project continuity,
and temporal conflicts, but you do not write executable actions or state_diffs.

## Goals
- Keep the session project-driven and grounded in the current household state.
- Use prior session narratives and carry-forward notes for temporal continuity.
- Choose plausible session_start_time/session_end_time and 2-4 time_segments.
- Create modest household variation: clarification, conflict, local observation,
  changed preference, forgotten device state, or rushed routine.
- Prefer executable actor actions for important state changes.
- Respect scene affordances: pick/place only pickable nodes; operate devices
  using supported state fields.

## Inputs
Session: {session}
Project: {project}
Current Scene: {scene_tree}
Memory Context: {memory_context}
Available Actors: {actor_ids}

## Output JSON
{
  "active_actor_ids": ["<persona_id>", "robot_001"],
  "session_start_time": "HH:MM",
  "session_end_time": "HH:MM",
  "time_segments": [
    {"segment_id": "seg_0", "start_time": "HH:MM",
     "end_time": "HH:MM", "summary": "..."}
  ],
  "session_script_summary": "What should happen, what the robot learns,
  and what concrete memory-relevant evidence should exist by the end."
}
\end{lstlisting}
\end{promptbox}
\caption{Condensed prompt excerpt for the Director Setup agent.}
\label{fig:director-setup-prompt}
\end{figure*}

\begin{figure*}[t]
\begin{promptbox}[Actor Turn --- System/User Prompt Excerpt]
\begin{lstlisting}[style=promptstyle]
## Role
You are {actor_name}, the in-world actor with id `{actor_id}`.
Follow the Director's instruction, but act naturally as this actor.
Output a short sequence of immediately executable event proposals.

## Rules
- Use only node IDs visible in the Action-Aware Scene Tree.
- Do not output state_diffs or hidden executor state.
- Use events[] only for this actor's next turn; usually 1-4 events.
- If unsure, ask a short clarification instead of inventing facts.
- Navigate before local actions; hold at most one object.
- Only pick/place targets marked pickable; do not move fixed furniture/devices.
- For set_device_state, use {"field": "<state name>", "value": <new value>}.
- Keep dialogue memory-useful: commitments, preferences, plans, object
  whereabouts, past recall, or physical/emotional state.

## Input
Your Context: {actor_context}

## Output JSON
{
  "events": [
    {"event_type": "utterance", "speaker": "{actor_id}", "text": "..."},
    {"event_type": "action", "actor_id": "{actor_id}",
     "action_type": "navigate_to|inspect|pick|place|open|close|set_device_state|handoff",
     "target_id": "<node id>", "args": {...}}
  ]
}
\end{lstlisting}
\end{promptbox}
\caption{Condensed prompt excerpt for actor turns. Proposed actions are validated by the executor before becoming trace events.}
\label{fig:actor-turn-prompt}
\end{figure*}

\begin{figure*}[t]
\begin{promptbox}[Session Examiner --- System/User Prompt Excerpt]
\begin{lstlisting}[style=promptstyle]
## Role
You are the Session Examiner for the EmbRealMem embodied-memory benchmark.
Read one just-finished agentic session and propose 0-3 memory QA candidates
that a future evaluator can ask to test long-term recall.

## Mission constraints
1. The question must require memory from this session and/or older sessions.
2. Prefer cross-session or cross-day evidence when past_session_briefs exist.
3. Every candidate must cite existing raw event_ids from session_events or
   past_session_briefs; never invent future or unseen evidence.
4. Use display names in query/reference_answer; raw ids appear only in evidence.
5. Ask one fact per question; avoid answer leakage in the query.
6. task_family is StateSingleHop, StateMultiHop, or TemporalMemory.
7. visibility_scope is robot_observed only if all evidence events include
   robot_001 in observer_ids; otherwise hidden_until_observed.

## Inputs
Current Session Header: {session_header}
Director Setup: {director_setup}
Node ID -> Display Name Map: {scene_brief}
Events of This Session: {session_events}
Past Session Briefs: {past_session_briefs}
Carry-Forward Notes: {carry_forward_notes}

## Output JSON
{
  "candidates": [{
    "task_family": "StateSingleHop|StateMultiHop|TemporalMemory",
    "sub_task": "ObjectLocation|DeviceState|RoutinePattern|Preference|TemporalCommitment|PartialObservation|InformationGap|ProjectContinuation|MultiHopReasoning",
    "query": "...",
    "reference_answer": "...",
    "acceptable_paraphrases": ["..."],
    "essential_memory_facts": ["..."],
    "evidence": [{"event_id": "...", "session_id": "..."}],
    "earliest_trigger": {"after_session_id": "...", "min_day_offset": 0},
    "visibility_scope": "robot_observed|hidden_until_observed",
    "difficulty": "easy|medium|hard",
    "answer_type": "location|state|entity|preference|time|count|yes_no|summary|other"
  }]
}
\end{lstlisting}
\end{promptbox}
\caption{Condensed prompt excerpt for the Session Examiner, which generates evidence-linked Memory QA candidates.}
\label{fig:session-examiner-prompt}
\end{figure*}

\subsection{Evaluation and ObsMem Runtime Prompts}
\label{sec:prompt-eval-obsmem}

We also expose the two prompts most relevant to evaluation transparency and
ObsMem reproducibility. Figure~\ref{fig:judge-prompt} shows the LLM-as-Judge
rubric used for Memory QA scoring, while
Figure~\ref{fig:obsmem-planner-selector-prompt} summarizes the ObsMem prompts
that convert a question into typed retrieval views and select final evidence.

\begin{figure*}[t]
\begin{promptbox}[LLM-as-Judge --- System/User Prompt Excerpt]
\begin{lstlisting}[style=promptstyle]
## Role
You are an evaluation judge for a household robot memory benchmark.
Evaluate how well the candidate answer uses information from the event
history to answer the question.

## Scoring criteria
0 - CONFLICT: the answer contradicts the reference or hallucinates.
1 - GENERIC: no contradiction, but the answer is generic or not memory-based.
2 - PARTIAL: the answer uses some relevant memory information.
3 - FULL: the answer uses all key information from the reference.

Treat acceptable paraphrases as equivalent to the reference answer.
Return JSON only:
{"score": <int 0-3>, "reason": "<brief explanation>"}

## User prompt fields
Question: {query}
Reference Answer: {reference}
Acceptable Paraphrases: {paraphrases}
Essential Memory Facts (judge-only): {facts}
Candidate Answer: {prediction}
\end{lstlisting}
\end{promptbox}
\caption{Condensed prompt excerpt for the LLM-as-Judge rubric used in Memory QA evaluation.}
\label{fig:judge-prompt}
\end{figure*}

\begin{figure*}[t]
\begin{promptbox}[ObsMem Query Planner and Evidence Selector --- Prompt Excerpt]
\begin{lstlisting}[style=promptstyle]
## Query Planner
Role: retrieval planner for a long-term memory system.
Task: convert the user's question into a compact JSON retrieval plan.
Do not answer the question. Prefer semantic interpretation over keywords.

Return JSON:
{
  "intent": "current_state|past_event|epistemic|commitment|planning|preference|temporal_pattern|open_qa",
  "answer_mode": "qa|planning",
  "target_entities": ["entities the answer depends on"],
  "related_entities": ["supporting participants, settings, or items"],
  "state_attributes": ["location", "holder", "power", "open", "timer", "..."],
  "temporal_filter": {"after": null, "before": null, "at": null,
                      "relative": "current|today|yesterday|future|past|any"},
  "evidence_views": ["state", "belief", "event", "reported", "commitment", "episode"],
  "rerank_strategy": "recency|semantic|hybrid",
  "action_goal": "<planning goal or empty>",
  "rationale": "<short explanation>"
}

## Evidence Selector
Role: select evidence for the same memory system.
Input: question, retrieval plan, and modality-tagged candidates.
Return an ordered list of the most informative candidates, up to {top_k}.
Prefer complementary evidence when candidates add new timestamps, entities,
or modalities. Prefer [state] for current-state questions, [belief] for
stale/uncertain states, [event]/[reported] for past events, [commitment] for
schedules or requests, and [episode] for broad cross-event questions.

Return JSON:
{
  "selected_indices": [candidate indices in priority order],
  "reasoning_chain": "<short description of how evidence answers the question>",
  "rationale": "<one short sentence>"
}
\end{lstlisting}
\end{promptbox}
\caption{Condensed prompt excerpt for ObsMem's query planner and evidence selector.}
\label{fig:obsmem-planner-selector-prompt}
\end{figure*}

\end{document}